\documentclass{article}
\usepackage{ijcai24}

\usepackage{graphicx}
\usepackage{times}
\usepackage{soul}
\usepackage{url}
\usepackage[hidelinks]{hyperref}
\usepackage[utf8]{inputenc}
\usepackage[small]{caption}
\usepackage[table]{xcolor}
\usepackage{amsmath}
\usepackage{amsthm}
\usepackage{booktabs}
\usepackage{algorithm}
\usepackage{algorithmic}
\usepackage[switch]{lineno}
\usepackage{xcolor}
\usepackage{pifont}
\usepackage{makecell}
\usepackage{array} 
\usepackage{tabularx}

\title{Neuromorphic Face Analysis: a Survey}

\author{
    Author Name
    \affiliations
    Affiliation
    \emails
    email@example.com
}

\iftrue
\author{
Federico Becattini$^{2*}$\and
Lorenzo Berlincioni$^{1*}$
\and
Luca Cultrera$^{1*}$\and
Alberto Del Bimbo$^1$\\
\affiliations
$^1$University of Florence\\
$^2$University of Siena\\
$^*$Equal contribution
\emails
$^1$\{name.surname\}@unifi.it,
$^2$\{name.surname\}@unisi.it
}
\fi

\begin{document}

\maketitle

\begin{abstract}
Neuromorphic sensors, also known as event cameras, are a class of imaging devices mimicking the function of biological visual 
systems. Unlike traditional frame-based cameras, which capture fixed 
images at discrete intervals, neuromorphic sensors continuously generate 
events that represent changes in light intensity or motion in the visual 
field with high temporal resolution and low latency.
These properties have proven to be interesting in modeling human faces, both from an effectiveness and a privacy-preserving point of view.
Neuromorphic face analysis however is still a raw and unstructured field of research, with several attempts at addressing different tasks with no clear standard or benchmark.
This survey paper presents a comprehensive overview of capabilities, 
challenges and emerging applications in the domain of neuromorphic face analysis, to outline promising directions and open issues.
After discussing the fundamental working principles of neuromorphic vision and presenting an in-depth overview of the related research, we explore the current state of available data, standard data representations, emerging challenges, and limitations that require further investigation.
This paper aims to highlight the recent process in this evolving field to provide to both experienced and newly come researchers an all-encompassing analysis of the state of the art along with its problems and shortcomings.
% Our findings indicate that neuromorphic sensors represent a
% highly promising field of computer vision that requires continuous 
% innovation to meet the growing demand for faster and more energy-efficient
% computing solutions with ultra-low latency.
\end{abstract}

\section{Introduction}
\label{sec:intro}

% In ordine sparso:
% \begin{itemize}
%     \item NeuroMorphic Camera, Definition.
%     \item Differences wrt to usual sensors.
%     \item Roba su privacy (FEDE)
%     \item Applicazioni
%     \item Differenze da RGB, vantaggi e svantaggi. Esempi dove l'event camera funziona meglio dell'approccio RGB.
%     \item challenge
%     \item datasets(CIONI)
%     \item trends
%     \item applicazioni future attese
%     \item cosa manca in letteratura
%     \item \todo{TODO: inventarsi qualche tabella (potrebbe avre senso una tabella con metodo, task di face-anlysis, tipo di dati (simulati-reali, anno)???) Oppure tabella con tutti i dataset in letteratura ???}
% \end{itemize}

Face analysis for decades has been one of the most studied topics in computer vision. Some tasks involving faces can even be considered to be solved, as they can be performed effectively in unconstrained scenarios with off-the-shelf tools. Face detection is likely the most spread-out application, as it is performed effortlessly by any kind of device, including personal smartphones.
A wide plethora of methods falls under the umbrella of face analysis: landmark detection, age estimation, lip reading, eye tracking, 3D reconstruction, just to name a few.
An interesting application of face analysis is the theoretical possibility of estimating human emotions and feelings just by observing facial micro-expressions \cite{ben2021video}. In fact, micro-movements of the face, triggered by fine muscle movements, have been mapped directly into emotions and it is known from psychology studies that such movements can be involuntary and almost impossible to hide \cite{ekman1978facial}. To this day, several computer vision applications that strive to estimate emotions by analyzing faces have been proposed, yet such micro-movements can happen at an extremely fast rate \cite{yan2013fast}, that is likely not going to be fully observable with a traditional RGB camera.
On the other hand, faces are arguably the most sensitive biometric data. Analyzing faces has thus raised privacy-related concerns, that have also been addressed in the recent AI Act by the European Commission \footnote{\url{https://www.europarl.europa.eu/RegData/etudes/BRIE/2021/698792/EPRS_BRI(2021)698792_EN.pdf}}.

These issues have recently led to an increasing interest in the usage of neuromorphic cameras, as they have shown promising results from different points of view, including effectiveness, latency, power consumption and privacy-preservation.
%Neuromorphic cameras represent a paradigm shift in the realm of visual sensing, drawing inspiration from biological vision systems.
Unlike conventional cameras that capture entire frames at fixed intervals, neuromorphic cameras operate on a fundamentally different principle, mimicking the asynchronous and event-driven nature of biological vision.
%The operational mechanism of neuromorphic cameras can be elucidated by the concept of "event-driven sensing".
%Unlike conventional cameras that capture and transmit entire frames regardless of scene changes,
Events are generated only when pixel-level changes in luminance exceed a predefined threshold. This approach enables the efficient use of computational resources, as only relevant information is transmitted and processed. The absence of a fixed frame rate means that these cameras can capture and process events with microsecond precision, a capability that is especially advantageous in dynamic and fast-paced environments.
This novel paradigm has the potential to open up new approaches for various applications, ranging from robotics and autonomous vehicles to surveillance and artificial intelligence.

% The asynchronous nature of event-driven sensing allows neuromorphic cameras to achieve unprecedented temporal resolution. The absence of a fixed frame rate means that these cameras can capture and process events with microsecond precision, a capability that is especially advantageous in dynamic and fast-paced environments.}

In this paper, we propose an overview of the relatively recent field of research involving event cameras and face analysis, which we dub Neuromorphic Face Analysis. Our goal is to provide a compendium for computer vision researchers in the field of face analysis, discussing opportunities and challenges, as well as taking stock of what is possible with event cameras up to this day. Not many works exist on the subject, yet they address several topics of primary importance and we firmly believe that neuromorphic face analysis will quickly raise the bar compared to traditional frame-based face analysis.

We first provide a brief overview of what an event camera is and how it differs from a traditional RGB camera and we discuss its advantages and challenges in the field of facial analysis in Sec. \ref{sec:adv_and_chal}. We also stress the importance of different event representation strategies and the advantages that event cameras can offer from a privacy-preservation point of view. We pass to discussing the applications and the state of the art (Sec. \ref{sec:research}), organizing the present research into macro-areas. Finally, we provide an analysis on the datasets available the literature in Sec. \ref{sec:datasets} and we discuss conclusions and future directions in Sec. \ref{sec:conclusions}.

\begin{table*}[t]

\renewcommand{\arraystretch}{2.5} % Adjust the value as needed

\resizebox{\textwidth}{!}{
\begin{tabular}{ccccccc}

%\cline{2-7}
  & \multicolumn{1}{c}{\makecell{\textbf{Face} \\ \textbf{Detection}}} & \multicolumn{1}{c}{\makecell{\textbf{Eye Blink} \\ \textbf{Detection}}} & \multicolumn{1}{c}{\makecell{\textbf{Pupil/Eye} \\ \textbf{Detection}}} & \multicolumn{1}{c}{\makecell{\textbf{Landmark} \\ \textbf{Detection}}} & \multicolumn{1}{c}{\makecell{\textbf{Drowsiness} \\ \textbf{Detection}}} & \multicolumn{1}{c}{\textbf{End-to-End}} \\ \hline
 \makecell{\textbf{Face} \\ \textbf{Detection}} &- &  \makecell{\cite{lenz2020event} \\  \cite{ryan2021real}} &  \cite{ryan2021real} &- &- &  \makecell{ \cite{bissarinova2023faces}  \\  \cite{ryan2021real}  \\  \cite{barua2016direct}}\\ \hline
 \makecell{\textbf{Identity} \\ \textbf{Recognition}} &  \cite{moreira2022neuromorphic} &  \cite{chen2020neurobiometric} &- &- &- &- \\ \hline
 \makecell{\textbf{Lip} \\ \textbf{Reading}} &  \makecell{ \cite{kanamaru2023isolated}  \\  \cite{tan2022multi}  \\  \cite{li2019lip}} &- &- & \cite{yu2022multimodal} & - &  \makecell{ \cite{bulzomi2023end}  \\  \cite{yoo2023rn}   \\  \cite{rios2023lipsfus}}\\ \hline
\makecell{\textbf{Voice Activity} \\ \textbf{Detection}} &  \cite{savran2018energy} &- &- & \cite{savran2023fully} &  - &- \\ \hline
\makecell{\textbf{Driver Monitoring} \\ \textbf{Systems}} &  \makecell{ \cite{liu2022neurodfd}  \\  \cite{ryan2021real}  \\  \cite{ryan2023real}} &  \cite{ryan2021real} &  \cite{ryan2021real} & \cite{liu2022neurodfd} &  \makecell{\cite{kielty2023neuromorphic} \\ \cite{chen2020eddd}} &  \makecell{ \cite{yang2022event}  \\  \cite{shariff2023neuromorphic}}\\ \hline
%\makecell{\textbf{Micro-Expr.} \\ \textbf{Recognition}} &  \cite{becattini2022understanding} &- &- &- &- &  \cite{guo2023gleffn}\\ \hline
\makecell{\textbf{Emotion/Expr.} \\ \textbf{Recognition}} &  \makecell{\cite{becattini2022understanding} \\ \cite{berlincioni2023neuromorphic}} &- &- & \cite{berlincioni2023neuromorphic} &  - & \cite{guo2023gleffn} \\ \hline
\makecell{\textbf{Gaze} \\ \textbf{Analysis}} &  \cite{ryan2023real} &- &  \cite{angelopoulos2020event} &- &- &  \cite{banerjee2022gaze}\\ \hline
\textbf{VR/AR} &  \cite{kang2023exploring} &- &  \cite{kang2023exploring} & \cite{kang2023exploring} &  - &- \\ \hline
\makecell{\textbf{Face Pose} \\ \textbf{alignment}} &  \cite{ryan2023real} &- &- &- &- &  \makecell{ \cite{savran2023multi}  \\  \cite{savran2023comparison}  \\  \cite{savran2020face}} \\ \hline

%\textbf{HCI} &- &- &- &- &- &\\ \hline

\end{tabular}
}
\caption{State of the art divided into intermediate modules (columns) and applications (rows).}
\label{tab:research}
\end{table*}

\section{Event Camera and Face Analysis: Advantages and Challenges}
\label{sec:adv_and_chal}
Event cameras exhibit unparalleled advantages in face analysis applications, offering a paradigm shift in capturing and interpreting facial dynamics. Their low-latency operation and high temporal resolution ensure real-time responsiveness to facial expressions and lip or eye movements, which are critical for applications like human-computer interaction and security systems.
%The high temporal resolution of event cameras enables the detailed tracking of dynamic facial changes, facilitating the nuanced analysis of subtle expressions or swift movements.
This capability or tracking subtle and swift dynamic facial changes proves invaluable in scenarios where traditional cameras may miss rapid facial transitions. Furthermore, the wide dynamic range of event cameras allows for accurate representation of facial features in challenging lighting conditions, offering improved performance compared to conventional cameras that may struggle with overexposure or underexposure.
%In essence, the advantages of event cameras in face analysis lie in their ability to capture and process facial dynamics with remarkable precision, making them particularly promising for applications demanding real-time, high-fidelity facial recognition and tracking. 
However, these advantages come with challenges. 
The asynchronous nature of event data, beneficial for real-time responsiveness, becomes a challenge in the field of face analysis: developing algorithms tailored to interpret facial expressions and movements from sporadic events poses a unique computational hurdle. Processing data produced by these sensors requires innovative approaches in computer vision to precisely recognize and understand facial dynamics.

It is crucial to note also that existing methods commonly employed for face analysis, such as face detectors and landmark detectors designed for traditional frame-based cameras, may not be directly suitable for event cameras. In Fig.~\ref{fig:outputs_vision} we show the outputs of a commonly used object detector and landmark detector (respectively DLIB \cite{dlib09} and Face Alignment \cite{bulat2017far}) trained on RGB frames from traditional cameras. We tested such models on video sequences captured with a paired set of event and RGB cameras. The object detector can properly locate targets only when the face motion is enough to make it visible (odd columns of Fig. \ref{fig:outputs_vision}). However, the outputs have an extremely low confidence due to the domain shift. The landmark detector instead fails to produce meaningful outputs, even if the face is fully visible (even columns of Fig.\ref{fig:outputs_vision}).
The asynchronous nature of event data and the lack of continuous frames present a mismatch with the assumptions underlying traditional face analysis methods. 
Furthermore, the limited availability of standardized datasets specifically designed for training event-based face analysis models is a notable impediment. Even how events are represented can pose a challenge, since models trained with a given representation are likely to lose effectiveness when such representation is changed.

\newcommand{\detwidth}{.24\linewidth}
\begin{figure}[t]
    \centering
    \includegraphics[width=\detwidth]{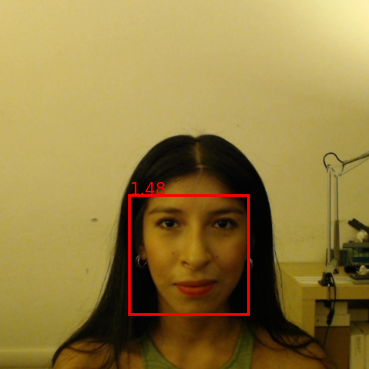}
    \includegraphics[width=\detwidth]{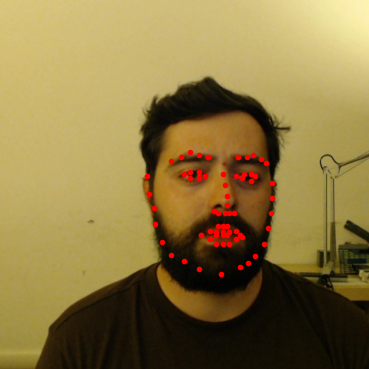}
     \includegraphics[width=\detwidth]{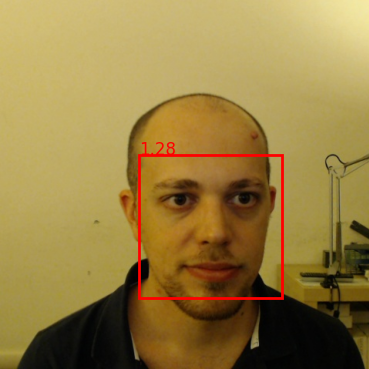}
      \includegraphics[width=\detwidth]{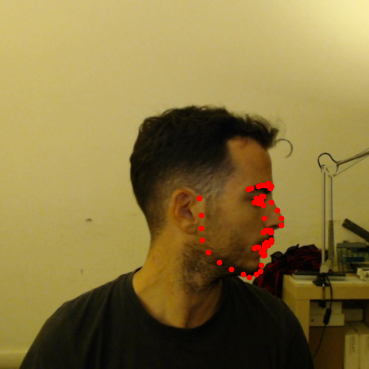}
      
    \includegraphics[width=\detwidth]{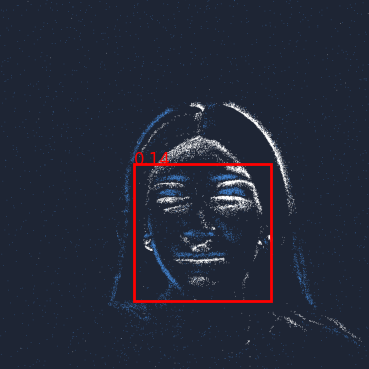}
    \includegraphics[width=\detwidth]{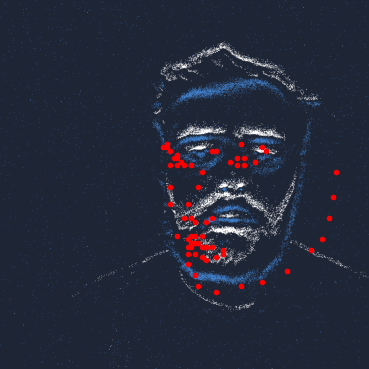}
     \includegraphics[width=\detwidth]{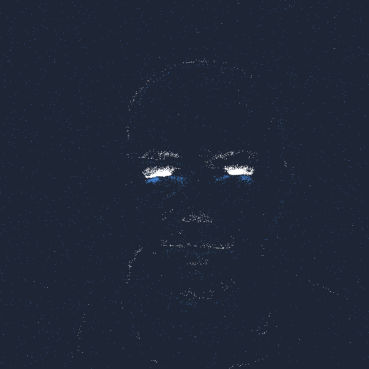}
     \includegraphics[width=\detwidth]{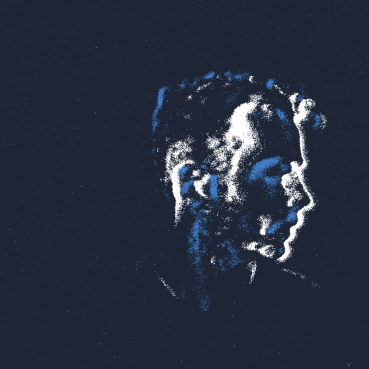}
     \caption{Failure cases of computer vision models over event data. \textit{Top Row}: Samples of face detection and landmark estimation on RGB frames.
    \textit{Bottom Row}: Samples of face detection and landmark estimation on the corresponding Event frames.}
         \label{fig:outputs_vision}
\end{figure}

%Calibration, essential for accurate facial feature mapping, is another critical challenge in event-based face analysis. Traditional calibration methods may not seamlessly adapt to the asynchronous data stream, necessitating the development of novel techniques to ensure the precise alignment of facial features with event camera pixels. This becomes crucial for achieving reliable and accurate face analysis outcomes, particularly in applications demanding high precision.

%Robust and comprehensive datasets capturing diverse facial expressions, lighting conditions, and demographic variations are essential for training models capable of generalizing well across various scenarios. The absence of such datasets hinders the development and validation of effective algorithms tailored to the distinctive characteristics of event cameras within the context of face analysis. 

%\todo{TODO: Eventi sintetici come possibile soluzione ma artefatti di compressione con citazione }

% \todo{ACCUMULAZIONE}Compared to standard sensors that acquire the absolute brightness of the view, event-cameras record the change in brightness. This fundamental mechanism implies that no signal is produced for \textit{static scenes}. In \cite{kielty2023neuromorphic} the trade-off between accumulation trough a time window and accumulation trough an event window is analyzed as a viable solution to the lack of information coming from the sensor.

% Efforts to overcome these challenges are critical to harness the full potential of event cameras in advancing real-time, dynamic face analysis applications. 

\paragraph{Data Representation}
An important challenge when dealing with event data is the fact that learning end-to-end models from raw events would require processing a huge amount of information, as millions of events can be generated every second. This poses a significant difference between the neuromorphic and RGB domains. To bridge this gap, researchers often engineer frame-like representations so that event data can be fed to a computer vision model, such as a convolutional neural network.

At the time, there is no clear standard and each encoding strategy must be tuned with specific hyperparameters. One of the most important is usually the accumulation time $\Delta t$, which controls the temporal granularity at which the information is processed.
Based on the task, different accumulation times can yield extremely different results. For instance, a face detector would require a sufficiently large $\Delta t$ in order to capture enough information for the face to be visible; on the contrary, modeling facial micro-expressions requires a sufficiently small accumulation time, as fast movements might be lost among other unrelated events.

The most common representation strategies involve quantizing events into spatio-temporal histograms. However, events can be processed also as raw individual events \cite{yu2022multimodal}. For this family of approaches minimal to none pre-processing is done.
The data keeps its original format or is quantized in extremely small accumulation times to simulate a continuous signal. This strategy often relies on the use of Spiking Neural Networks \cite{yu2022multimodal,li2019lip}.
%$(x_i , y_i, t_i, p_i)$ (where $x_i, y_i$ are the coordinates of the event, $t_i$ the event timestamp, and $p_i$ its polarity).
%This more geometric, point-wise strategy chooses a raw uncompressed representation that need large amount of memory. 
Other works have adopted an intermediate strategy, i.e. do not use a frame-based representation yet increase the accumulation interval $\Delta t$, obtaining a voxel of quantized events in the form of a 3D tensor \cite{bulzomi2023end,yang2022event}.

To treat the task as a more standard computer vision problem, it is common to transform the raw data to a 2D grid (an \textit{image}). This allows the use of models originally developed for processing RGB data.
This accumulation can be done, for example, by counting events in a spatio-temporal neighborhood \cite{kielty2023neuromorphic}, treating sequences of binary events as binary digits of a decimal number \cite{innocenti2021temporal,becattini2022understanding} or building a motion history image by applying an exponential decay over the neighborhood of active events \cite{mueggler2017fast}.

\paragraph{Event Cameras as a Privacy Preservation Layer}
\label{sec:privacy}

A notable advantage of neuromorphic vision is that several works have attributed to the usage of an event camera the benefit of working under the preservation of privacy \cite{al2020privacy,delilovicbio,han2023towards,dong2023bullying10k}. Compared to RGB images, in fact, event streams offer an additional layer of protection, as they are harder to interpret and since no signal is generated when both the camera and the observed environment or subject are static.
In particular, \cite{al2020privacy} and \cite{dong2023bullying10k} leverage event cameras for action recognition purposes, while protecting the identity of the analyzed people. This idea is used by \cite{dong2023bullying10k} to work in sensitive environments such as high schools, where minors are filmed to prevent and recognize episodes of bullying.
Nonetheless, whereas it is certainly true that intensity information is discarded in event data, sensitive information might still be recoverable from events \cite{rebecq2019events}.
The authors of \cite{zhang2023event} have thus addressed the problem of encrypting events so to be safely transmitted over an untrusted channel. In particular, they proposed an encryption strategy that can prevent the usage of a computer vision model directly on the protected data.
Adopting an opposite paradigm, \cite{ahmad2022event} proposed an event scrambling method that makes the stream impossible to interpret for the human eye but retaining the possibility of applying computer vision models effectively, showing that it is even possible to perform person re-identification on protected data.
Finally, \cite{delilovicbio} and \cite{han2023towards} studied the advantages of processing events on the edge and on nodes of a federated network. What these works propose is to exploit the low power consumption of event cameras to enable an efficient computation directly on the edge, without the need to transmit the data in the first place. Despite proposing an edge computing paradigm, \cite{delilovicbio} also stresses the fact that event cameras provide an additional layer of privacy due to the lack of intensity information.

\section{Present Research}
\label{sec:research}
We now present an overview of the current literature in the field of event-based vision, related to human faces.
Neuromorphic Face Analysis has been applied to study several sub-topics in the recent literature. We identified a collection of low-level modules, that are commonly used as an intermediate step to address different applications. Low-level modules solve simpler tasks, that provide information about the face such as face detection, landmark detection, pupil detection, etc. The outputs of such modules can be leveraged for downstream applications, spanning from identity recognition to driving monitoring systems.
The only notable exception is face detection, which we consider to be both a low-level module and an application, as it is commonly framed in both declinations.
It is also worth noticing that, not all works in the literature rely on intermediate modules, tackling the problem in an end-to-end fashion.

We summarize in Tab.~\ref{tab:research} the most common modules and applications that have been addressed in the literature with an event camera so far, explicitly referring to the corresponding works from the state of the art.
As highlighted in Tab.~\ref{tab:research}, we identified nine macro areas in which faces are analyzed with an event camera. In the following, we provide a more in-depth analysis of these lines of research.

%We begin with the most common task, face detection, and then extend our exploration to more complex tasks directly stemming from face detection. These tasks include face pose alignment, lip reading, and the detection and tracking of gaze and pupil, introducing additional facets of visual perception.

%We then delve into research leveraging these advanced features in broader contexts, such as emotion and expression recognition. Here, a deeper understanding of facial dynamics can lead to significant advancements. Finally, we will examine how these technologies can be integrated into practical scenarios, such as assisted driving, where accurate perception of facial expressions and movements can greatly contribute to the interaction and communication between the autonomous system and the human user.
%This gradual progression will allow us to navigate the increasingly intricate landscape of event-based vision applications, specifically concerning human faces.

\paragraph{Face Detection}
As discussed in Sec.~\ref{sec:adv_and_chal}, face detection from event streams is not trivial. However, it is arguably the most important application in neuromorphic face analysis as it enables most face-related tasks. Some works in the literature, have developed neuromorphic face detectors \cite{lenz2020event,ryan2021real,bissarinova2023faces,barua2016direct}.

%\textbf{Face Detection \& Identity Recognition}
%Face detection is the pivotal task that automates the localization of human faces in images or videos. This foundational step not only lays the groundwork for understanding facial dynamics but also extends to identity recognition, enabling the unique association of faces with specific individuals. 
%Face detection with event cameras proves to be a non-trivial undertaking. Unlike traditional frame-based cameras, event cameras capture changes in luminance asynchronously, presenting unique challenges in adapting conventional face detection methodologies. Navigating the complexities of event-based vision demands innovative approaches to account for the dynamic nature of the data, making the task of identifying faces in this context notably intricate and requiring specialized methods.

\cite{barua2016direct}  was the first to address the problem of face detection from event streams. The approach is based on a patch-based model that analyzes small crops of intensity changes and determines whether they are likely to be caused by a face. The method was able to achieve good performance on a benchmark dataset of event streams, demonstrating the feasibility of face detection from event data. Starting from this pioneering work, \cite{lenz2020event}  introduced an alternative approach to face detection by utilizing eye blinks as a distinctive cue to identify the presence of a face within a scene.

Differently from these approaches, which specifically leveraged motion-based features derived from the neuromorphic streams, a few attempts addressed the task in a similar way to standard solutions in the RGB domain. \cite{ryan2021real}  introduced a gated recurrent YOLO (GR-YOLO) architecture for multi-face and eye detection. However, this work relied on Neuromorphic-Helen (N-Helen), a synthetically generated dataset, converted from RGB videos.
To attempt to bridge the gap between synthetic and real events, \cite{bissarinova2023faces} collected an annotated neuromorphic face detection dataset. In addition, they propose also 12 baseline methods, derived from the RGB literature to predict bounding boxes and facial landmark coordinates.

Other works have followed similar strategies to develop face detection modules finalized to solve a downstream task, such as identity recognition \cite{moreira2022neuromorphic}, emotion recognition \cite{berlincioni2023neuromorphic}, gaze analysis \cite{ryan2023real}, among others. We will discuss in-depth these methods in the following paragraphs.

\paragraph{Identity Recognition}
Identity recognition is a fundamental task in traditional computer vision, as it can be applied for surveillance and security. Neuromorphic vision offers a valuable asset in this direction, as its incredibly low latency can provide a rich characterization of biometric traits.
In fact, \cite{chen2020neurobiometric}  presented a biometric authentication system that utilized eye blinks as a unique and distinctive feature for verifying individuals. 
%These features capture the dynamic characteristics of eye blinks, which are highly individual. The features are then encoded into a fingerprint-like representation, which serves as the unique biometric identifier for the user performing identity recognition.

To quantify how much identity-related information is carried by facial movements and how such information can be extracted by an automatic neuromorphic system, \cite{moreira2022neuromorphic} proposed a different event-based approach for face identity recognition: the authors aggregated events in frames, that were then normalized and grouped into \textit{face tokens}. These tokens represent the facial activity in a specific time window and are analyzed with a spatio-temporal 3D Convolutional Neural Network (3DCNN). The research demonstrated the crucial role of facial dynamics in identity recognition and the viability of neuromorphic sensors in detecting subtle facial movements. It also presented NVSFD, a dataset tailored for speech-induced facial dynamics.

Despite these findings, this remains quite an unexplored field of research, as to the best of our knowledge no other work on the subject exists yet. Similar ideas, however, have been explored for event-based re-identification from full bodies \cite{ahmad2022event}.

% In \cite{rybski2004cameo} : The authors propose CAMEO (Camera Assisted Meeting Event Observer), that uses a combination of RGBs and event cameras to observe and understand meeting dynamics. 
%The traditional camera is used to capture high-resolution images of the meeting room, while the event camera is used to capture high-speed data about changes in light intensity.
% This combination of data allows the system to track the location and movements of people in the meeting room with high accuracy, even in low-light conditions or when people are partially occluded.
% The CAMEO system also performs face detection to identify individuals in the meeting room. This is done by using the event camera data to track the motion of faces and then using the traditional camera data to identify specific faces.

\paragraph{Face Pose Alignment}
A challenging application in the context of human-robot interaction concerns estimating face poses. This application can also serve as an important pre-processing step for face analysis tasks, as it determines the orientation and position of facial features.
A few approaches to estimate face pose alignment have been recently proposed.

%In the context of event-based vision, face pose alignment emerges as a critical facet, extending beyond conventional face detection to precisely determine the orientation and position of facial features in dynamic scenarios.
%This section explores related works dedicated to face pose alignment using event cameras.
%\todo{By examining existing research endeavors, we aim to unravel the innovative methodologies and approaches employed in achieving accurate face pose alignment within the asynchronous and dynamic framework of event-based vision. RIMUOVERE?}
\cite{savran2020face} were the first to address this task with event cameras. The alignment is achieved through a regression cascade of tree ensembles. Efficiency is also taken into account, as the alignment is initialized only when a pose change is detected to minimize unnecessary processing.
The authors also recorded a dataset with human subjects exhibiting large head rotations, varying movement speeds, speaking intervals, and multi-human annotations. %The dataset is made available upon request.

The same authors also extended this approach in \cite{savran2023multi} and \cite{savran2023comparison}, by enhancing it with a multi-timescale event encoding strategy.
In \cite{savran2023multi}  multiple timescales are used to improve the efficiency and quality of face pose alignment. This is achieved by adaptively adjusting the processing rate based on facial movement intensity. The effect is that the method generates sparse pose-events, reducing computational demands.
%This approach excels in achieving heightened accuracy without compromising on minimal delay, making it particularly well-suited for real-time applications.
Further analysis was carried out in \cite{savran2023comparison}, where two different timing strategies for face pose alignment are used: constant time frame and constant event count frame.
This work addressed the problem of determining the appropriate accumulation time intervals for the face alignment problem.
%that the constant event count strategy outperforms the constant time frame strategy in terms of face pose alignment performance and efficiency.

Face pose alignment has also been addressed by \cite{ryan2023real} for driver monitoring system applications.

\paragraph{LipReading \& Voice Activity Detection}
Given the capacity of event cameras to model high temporal resolution signals, several works have leveraged neuromorphic strategies to analyze mouth-related tasks, such as lip reading and Voice Activity Detection (VAD).
VAD is a method for identifying and isolating periods of speech within an audio stream; differently, lip-reading is the process of understanding spoken language by observing the movement of a person's lips.
In both tasks, the micro-movements of the mouth must be captured to identify specific utterances.

In the exploration of these tasks, researchers have adopted diverse approaches, some leveraging both audio and video modalities, while others focus exclusively on video inputs.

A pioneering approach focusing on the task of Voice Activity Detection using both audio and video was proposed in  \cite{savran2018energy}.
%This work introduced a method that utilizes data from both audio and event streams to detect speech.
The pipeline starts by jointly locating and detecting lip activity using a probabilistic estimation technique after applying spatio-temporal filtering. VAD is then performed, combining visual and audio features.
In a similar vein, \cite{savran2023fully} proposed an event intensity-based approach by constructing a fully convolutional network for effective neuromorphic VAD. The significance lies in the development of a purely vision-based VAD model, breaking away from traditional audio-centric approaches.

Also lip reading approaches have declined the task as a video or an audio-video approach. Among the video-based approaches, \cite{tan2022multi}  proposed MSTP, a Multi-grained Spatio-Temporal features Perceived network. The model comprises two branches processing low-rate and high-rate event frames: on the one hand, the low-rate branch accumulates event for a longer time-span, thus capturing complete spatial features; on the other hand, the high-rate branch focuses on fine-grained temporal features.
The two branches are connected throughout the network using message flow modules merging the features.
The authors also addressed the lack of event-based lip-reading datasets by creating DVS-Lip, a dataset with 19,871 word samples.

The idea of separating neuromorphic information in independent streams at different resolutions has been also followed in \cite{yoo2023rn}. The authors propose RN-Net, an architecture specifically designed to process asynchronous temporal data. It employs simple convolution layers seamlessly integrated with dynamic temporal encoding reservoirs to effectively detect spatio-temporal features at both local and global levels.
Reservoirs are composed of a large number of interconnected neurons, and they can be thought of as a kind of "memory" for the network.
Similarly, \cite{bulzomi2023end} and \cite{kanamaru2023isolated} have explored the use of event-based cameras for lip reading. Differently from prior work, \cite{bulzomi2023end} proposed a spiking neural network architecture for this task, demonstrating an improvement in accuracy. \cite{kanamaru2023isolated} instead leveraged a hybrid approach that combines event-based and frame-based camera data and used a Temporal Convolutional Network to recognize sounds.
They also proposed an original dataset of 15 Japanese consonants from 20 speakers.

Differently from VAD, the idea of combining audio with event cameras in the field of lip reading is more recent. Among these works,
%Several works have instead adopted a multimodal approach, leveraging both video and audio signals.
\cite{yu2022multimodal}  introduced a spiking neural network approach for audio-visual speech recognition based on lip reading. Their innovative use of liquid state machines and a soft fusion method inspired by the attention mechanism showcases the effectiveness of a combined audio-visual approach also for the lip reading task.

Interestingly, the neuromorphic paradigm has been adopted for both the video and audio modalities. A few works have in fact used neuromorphic audio sensors along with neuromorphic cameras \cite{rios2023lipsfus,li2019lip}.
\cite{li2019lip}  presented a multi-modal fusion deep network for event-based lip reading using spiking sensors, combining Dynamic Video Sensors (DVS) and Dynamic Audio Sensors (DAS). The fusion model enhances lip reading performance by integrating visual and auditory information, reflecting the multimodal nature of the approach.
To address the shortage of labeled data, \cite{rios2023lipsfus}  introduced the LIPSFUS dataset, which comprises both visual and auditory information. The visual information consists of the lip movement when a subject articulates a word, while the auditory information refers to the sound that a subject makes when pronouncing a word.
The data has been captured by a Neuromorphic Auditory Sensor (NAS) and a DVS camera and is synchronized with the same timing source.
%Lip reading, on the other hand, is likely to represent a greater challenge than VAD, especially in the event context. Indeed, VAD is less affected by noise or occlusion as lip reading is, as it often relies also on audio.

\paragraph{Facial Expression \& Emotion Recognition}
One of the most challenging, yet suitable for event-based approaches, is the analysis of facial expressions and their underlying emotions. The difficulty of these tasks is due to the extremely high temporal resolution at which facial micro-movements happen to express an emotion. This poses a natural limit for standard RGB approaches, which makes the usage of neuromorphic vision sensors an interesting solution.

In this context, Becattini et al. \cite{becattini2022understanding}  proposed a pioneering method for modeling expressions with event cameras. The method employs an event-based convolutional neural network that analyzes temporal patterns of brightness changes to identify micro-expressions. The study demonstrated that, by using event cameras, it is possible to understand human reactions solely by observing facial expressions. A comparison with RGB-based modeling showed the improved effectiveness of the approach. However, this work only detected positive, neutral or negative reactions. This idea was then extended in \cite{berlincioni2023neuromorphic}, where a dataset for Neuromorphic Event-based Facial Expression Recognition (NEFER) was introduced, containing annotations for the 7 basic emotions. The dataset consists of paired visual spectrum images and event camera streams. For each sequence of frames, both the expected emotion and the self-reported emotion from the user are provided, allowing for a holistic understanding of emotional expression. They also proposed a baseline method using a 3D convolutional network for emotion recognition.

A more fine-grained analysis was recently proposed in \cite{guo2023gleffn}. The authors developed an approach to micro-expression recognition called GLEFFN. Two main components are present: an event-based feature extraction module and a global-local feature fusion network.
The event-based feature extraction module extracts a local count image from an up-sampled video using SloMo \cite{jiang2018super}. The local count image is a binary image where each pixel is either on (indicating an event) or off (indicating no event). 
The global-local feature fusion network combines the local count image with global dense optical flow to obtain deeper features. By fusing the local and global features, the network can capture both local and global motion information, which is crucial for accurate micro-expression recognition.
The problem of facial expression recognition has also been declined as a valence-arousal estimation problem in \cite{berlincioni2024neuromorphic}.

\paragraph{Gaze Analysis and AR/VR}
Another intriguing application of event cameras is gaze and pupil detection and tracking. Event cameras are well-suited for this task due to their high temporal resolution, which allows them to accurately track the movements of the eye. This application is closely related to augmented and virtual reality scenarios, as gaze is a natural way of interacting with head-mounted devices.

The challenge behind gaze analysis lies in the fact that eye saccades can happen abruptly and very quickly. \cite{angelopoulos2020event}  presented a hybrid frame-event near-eye gaze tracking system that delivers high update rates up to 10,000~Hz. Their system maintained an accuracy comparable to high-end desktop-mounted commercial trackers. The system utilizes an event camera that captures both high-frequency events and low-frequency frames to provide a more efficient and accurate representation of eye motion.
The authors introduced a model-based eye tracking algorithm that operates at the event rate and an online 2D pupil fitting method that updates a parametric model every one or few events. The system also employed a polynomial regressor to estimate the point of gaze from the parametric pupil model in real-time.
%This work paves the way for a new generation of ultra-low-latency gaze-contingent rendering and display techniques for virtual and augmented reality.

Differently, \cite{banerjee2022gaze} presented an event-based gaze detection system using retinomorphic events recorded on a DVS camera. The authors developed a new and compact event-based dataset for gaze detection under various lighting conditions. They also proposed a novel event encoding technique that encodes event logs into six-channel images and then used a Convolutional Neural Network for gaze prediction. The authors evaluated their method with multiple metrics and found that it achieved high accuracy in gaze prediction.
Finally, \cite{kang2023exploring}  developed a remote pupil-tracking technique using event cameras for Autostereoscopic 3D displays and AR 3D Head-up displays (HUDs). %\todo{CONTROLLARE IL PAPER: The authors also discuss the challenges and limits of using event camera imaging for remote pupil tracking.}

\paragraph{Driver Monitoring System}
Event cameras have proven to be attractive also for developing Driver Monitoring Systems (DMS). A DMS is a tool that is used to improve safety while driving as it can detect fatigue, distractions, or altered states of the driver.
This application shares some traits with all the aforementioned applications since fatigue and distraction indicators can be tied to facial expressions and movements. Basic modules such as face detectors, landmark estimators or pupil trackers find a natural applicability in the development of such applications.

A driving monitoring system can address different issues.
The most basic problem is to localize the driver in the frame. \cite{liu2022neurodfd}  proposed a driver face detection framework called NeuroDFD that uses event-based data. The proposed method involves constructing event representation, incorporating a shift feature pyramid network and shift context modules that process temporal information at different scales. This approach also yielded 5 facial landmarks and posed itself as a preliminary step for developing more complex monitoring systems.
Similarly, \cite{ryan2023real} and \cite{ryan2021real} also detected the faces of drivers. \cite{ryan2023real} developed a face detection module jointly estimating head pose, eye gaze and facial occlusions in real-time.
The framework is trained on synthetic event-stream data from conventional video datasets and validated on real event camera data.
\cite{ryan2021real} instead predicted eye blinks and pupil detections in addition to facial bounding boxes.

A different take on the problem has been proposed by \cite{kielty2023neuromorphic} and \cite{chen2020eddd}, who proposed to directly infer the drowsiness state of the driver.
\cite{kielty2023neuromorphic} use event-based data to analyze mouth movements in search of yawning behaviors that provide a complementary indicator of tiredness. They also proposed a dataset of 952 video clips recorded with an RGB camera and converted into neuromorphic images using a video-to-event converter \cite{Hu2021v2ecvprworkshopeventvision2021}.
Instead of only detecting yawns, \cite{chen2020eddd} proposed to also recognize eye blinks and mouth movements. Additionally, they provided the EDDD dataset, dedicated to event-based drowsiness driving detection. 
Finally, \cite{shariff2023neuromorphic} analyzed faces of drivers to estimate distraction.
The authors developed a sparse-ResNet, that extracts features efficiently from event data and classifies the driver's distraction. Furthermore, two synthetic event datasets, Drive\&Act and DMD, were created to train and evaluate the proposed model.
Interestingly, the same task has also been addressed by looking at both faces and bodies in \cite{yang2022event}.
The paper proposed a real-time driver distraction and action recognition. The performance of the proposed system was evaluated on a large-scale simulated event dataset and a self-recorded real event dataset with a DAVIS346 event camera. Additionally, the authors implemented transfer learning experiments on real event data, demonstrating promising generalization capabilities.

\begin{table}[t]
\centering
    \resizebox{\linewidth}{!}{
		\begin{tabular}{lccccl}
         \textbf{Dataset} & \textbf{Video}  & \textbf{Users}  & \textbf{Resolution} & \textbf{Public} & \textbf{Annotations}  \\
         \hline
         \cite{savran2020face} & 108  & 30 & 304$\times$204 & X & Eye; Lip\\
         \cite{lenz2020event} & 48 & 10 & 640$\times$480 & \ding{51} & Eye blink \\
         \cite{becattini2022understanding} & 455 & 25 & 640$\times$480 & X & Binary Reactions \\
         \cite{berlincioni2023neuromorphic} & 609 & 29 & 1280$\times$720 & \ding{51} & Face; Landmk.; Emotion \\
         \cite{bissarinova2023faces} & 3889 & 73 & 408$\times$360 & \ding{51} & Face; Landmk. \\
         \cite{tan2022multi}  & 200 & 40 & 346$\times$260 & \ding{51} & Lip; Utterances \\
         \cite{chen2020eddd} & 260 & 26 & 346$\times$260 & \ding{51} & Voice; Eye blink; Yawn\\
         \cite{banerjee2022gaze} & 3360 & 6 & 1920$\times$1080 & \ding{51} & Pupil Coordinates \\
         \cite{angelopoulos2020event} & 24 & 24 & 346$\times$260 & \ding{51} & Gaze \\
         \cite{rios2023lipsfus} & 22 & 110 & 128$\times$128 & \ding{51} & Utterances \\
         \cite{barua2016direct} & - & 30 & 128$\times$128  & X & Face\\
         \cite{moreira2022neuromorphic} & 436 & 40 & - & \ding{51} & Identity \\
         \cite{kanamaru2023isolated} & 1500 & 20 & 1280$\times$800 & X & Face; Landmk.; Lip \\
         \cite{ryan2023real} & - & 5 & 1280$\times$720 & X & Head pose \\
         \cite{chen2020neurobiometric} & 180 & 45 & 346$\times$260 & \ding{51} & Eye blink \\
         \cite{savran2018energy} & 360 & 18 & 304 $\times$ 240 & X & Face; Lip; Voice \\
         
    \end{tabular}
    }
    \caption{Several dataset focused on human faces expression and detection annotated with their respective attributes.}
    \label{tab:datasets}
\end{table}
\section{Datasets}
\label{sec:datasets}
At the time of writing, there are no well-established benchmarks and the lack of event camera data poses a challenge to the development of new face analysis models.
However, several works that analyze faces with neuromorphic cameras collected footage for carrying out their experiments. Tab. \ref{tab:datasets} provides an overview of the datasets that have been collected in the past years for neuromorphic face analysis. We would like to stress the fact that most of these datasets contain a small amount of samples with a lack of diversity. At the same time, the spatial resolution of these datasets is limited due to hardware limitations, since only in the last few years event cameras have been equipped with Full HD sensors (1920 $\times$ 1080).
We report in Tab. \ref{tab:datasets} also the annotations that are made available along with the recorded videos.
These annotations include low-level labels such as bounding boxes of faces (\texttt{Face}), lip regions (\texttt{Lip}), eye regions (\texttt{Eye}) and coordinates of specific facial keypoints~(\texttt{Landmk.}). Some datasets also include higher-level annotations such as positive/negative facial reactions to visual stimuli (\texttt{Binary reactions}), words pronounced by the subjects (\texttt{Utterances}) and whether a subject is speaking (\texttt{Voice}).

It is worth noticing that annotating events is not straightforward. First, instead of annotating frames, the labels must be related to a set of spatio-temporal events. Then, since an event camera does not produce any signal in the absence of motion, identifying the entities to annotate is harder. For instance, one may want to annotate a static face, even when it is not visible. In addition, events are often accumulated to be processed with a frame-based model. The accumulation time and the encoding strategy directly affect the quality or the granularity of the annotations.

\paragraph{Synthethic Approaches}
\label{synth_data}
It must be noticed that all dataset in Tab. \ref{tab:datasets} are made of real event data, i.e. recorded with an actual neuromorphic sensor. However, a large crop of literature \cite{guo2023gleffn,kielty2023neuromorphic,ryan2021real,ryan2023real,savran2020face,yu2022multimodal,berlincioni2024neuromorphic} works with synthetic events, obtained by converting RGB videos into neuromorphic streams. Also some of the methods in Tab. \ref{tab:datasets} \cite{barua2016direct,becattini2022understanding,chen2020eddd,shariff2023neuromorphic} leverage synthetic events to perform additional experiments or train on more data. Some works, such as \cite{becattini2022understanding} also convert RGB videos to event streams to trivially annotate events by transferring any available label attached to the original data.

%Due to the aforementioned difficulties of collecting real-world event-data multiple researchers focused on using simulators to obtain the neuromorphic equivalent of the easily available RGB videos.
Different event camera simulators have been proposed in the literature, namely, ESIM \cite{Rebecq18corl} and V2E \cite{Hu2021v2ecvprworkshopeventvision2021}. These simulators are capable of producing neuromorphic counterparts from RGB videos. To this end, they first perform a temporal upsampling of RGB frames, with a rate that adapts to the video content and its estimated visual dynamics (the more the video changes, the more frames are added). Then, synthetic events are generated by analyzing the differences between adjacent frames. 

\begin{figure}[t]%
\centering
\includegraphics[width=0.49\columnwidth]{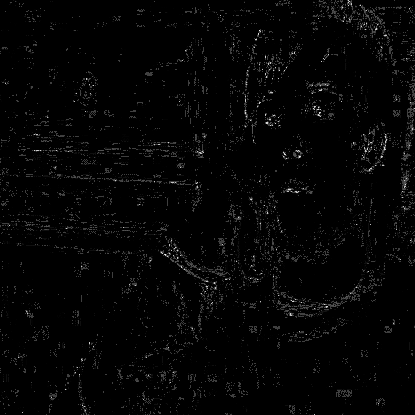}
\includegraphics[width=0.49\columnwidth]{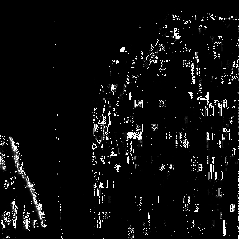}
\caption{Compression artifacts after simulation using  \protect\cite{Hu2021v2ecvprworkshopeventvision2021}. Image courtesy of \protect\cite{berlincioni2024neuromorphic}}
\label{fig:compressed1}
\end{figure}

While these tools have proven to be beneficial for many tasks they also present drawbacks.
As discussed in \cite{berlincioni2024neuromorphic}, from which we report also here Fig. \ref{fig:compressed1},
%and Fig.\ref{fig:compressed2},
the RGB to Event simulators are sensible to the compression of the source video, to the point that the block coding artifacts, barely visible in the original footage, are exaggerated and turned into block-sized \textit{macro}-events. This hinders the applicability of simulators and underlines the domain shift between real and simulated events.

% \begin{figure}[t]%
% \centering
% \includegraphics[width=0.32\columnwidth]{imgs/hi_res.png}
% \includegraphics[width=0.32\columnwidth]{imgs/crf30.png}
% \includegraphics[width=0.32\columnwidth]{imgs/crf40.png}
% \caption{Compression artifacts in progressively more compressed source videos getting simulated using \protect \cite{Hu2021v2ecvprworkshopeventvision2021}.  }
% \label{fig:compressed2}
% \end{figure}

%\section{Discussion}

% \section{Guide}
% \todo{TODO: Trovare titolo. \\Neuromorphic face analysis. A survey of}
% % \begin{itemize}
% %     \item Face Detection
% %     \item Landmark Detection
% %     \item Emotion Recognition
% %     \item Pupil/Gaze Detection (?)
% % \end{itemize}

% \begin{itemize}
%     \item FIG(S) ?!
%     \item Tanti dati con moto
%     \item Nessun segnale se stai fermo
%     \item Codifica
%     \item Tempo di accumulazione
% \end{itemize}

% \todo{TODO: del testo utile non generato.
% As previously stated before these sensors introduce a new paradigm along with several challenges. Compared to standard sensors that acquire the absolute brightness of the view, event-cameras record the change in brightness. This fundamental mechanism implies that no signal is produced for \textit{static scenes}. In \cite{kielty2023neuromorphic} the trade-off between accumulation trough a time window and accumulation trough an event window is analyzed as a viable solution to the lack of
% }

\section{Conclusions and Future Developments}
\label{sec:conclusions}

Neuromorphic Face Analysis is a field of research at a very early stage. Nonetheless, several works have underlined the effectiveness of neuromorphic cameras for capturing facial features, offering several benefits compared to traditional computer vision. First of all the extremely low latency allows to capture facial micro-movements which might not be fully observable with RGB cameras. The high dynamic range of the sensors is an aid for stable driver monitoring systems as well as person identification modules under challenging conditions.
In this paper, we have discussed present research on the subject, pointing out that there are still many challenges to be addressed. The lack of annotated data poses a big issue: no standard benchmark exists and new datasets must be collected even for addressing tasks that are close to being solved in RGB vision, such as face detection. Many works have bypassed this problem, relying on synthetic data obtained with a simulator. However, whereas this could be a suitable solution for many problems, we believe that real events should be leveraged to exploit the full capacity of neuromorphic sensors for analyzing faces as micro-movements are lost when converting low-framerate RGB videos.
In particular, we found the field of expression and emotion recognition to be almost unexplored. This comes as a surprise as we believe that observing the extremely fast movements that convey emotions is a perfect application for a sensor with such a low latency as an event camera. Studying emotions would also allow the development of applications in healthcare and human-computer interaction. 
For instance, in the healthcare field, the development of neuromorphic medical devices could monitor the patient's pain expression.

Neuromorphic sensors could also play a crucial role in advancing augmented reality technologies. Whereas a few works in this direction are present in the literature, these sensors could enable more immersive and responsive AR experiences, enhancing how users interact with digital information overlaid on the real world.
The energy-efficient nature of neuromorphic sensors makes them also well-suited for edge computing applications. Future developments may see the integration of these sensors into small, resource-constrained devices, contributing to the growth of decentralized and efficient computing.

In conclusion, the ongoing research and innovations in this area are likely to shape the next generation of intelligent systems and contribute to the continued advancement of artificial intelligence.

\begin{small}
\textbf{Acknowledgments}
This work was supported by the European Commission
under European Horizon 2020 Programme, grant number
951911—AI4Media.
This work was partially supported by the Piano per lo Sviluppo della Ricerca (PSR 2023) of the University of Siena - project FEATHER: Forecasting and Estimation of Actions and Trajectories for Human-robot intERactions.
\end{small}

\bibliographystyle{named}
\bibliography{ijcai24}

%\clearpage

\end{document}